\documentclass[journal]{IEEEtran}

\usepackage{subfigure}
\usepackage{times}
\usepackage{epsfig}
\usepackage{graphicx}
\usepackage{amsmath}
\usepackage{amssymb}
\usepackage{multirow}
\usepackage{color}
\usepackage{cite}
\usepackage{array}

\begin{document}

%\title{Deep Hashing Learning Simultaneously for Remote Sensing Image Retrieval and Classification}
%
\title{Deep Hashing Learning for Visual and Semantic Retrieval of Remote Sensing Images}

\author{Weiwei Song,~\IEEEmembership{Student Member,~IEEE,} Shutao Li,~\IEEEmembership{Fellow,~IEEE,} and J\'on Atli Benediktsson,~\IEEEmembership{Fellow,~IEEE}

%\thanks{This work was supported by the National Natural Science Fund of China under Grant 61890962 and 61520106001, the Science and Technology Plan Project Fund of Hunan Province under Grant CX2018B171, 2017RS3024, 2018TP1013, and the Science and Technology Talents Program of Hunan Association for Science and Technology under Grant 2017TJ-Q09.}

\thanks{W. Song and S. Li are with the College of Electrical and Information Engineering, Hunan University, Changsha 410082, China, and also with the Key Laboratory of Visual Perception and Artificial Intelligence of Hunan Province, Changsha 410082, China (e-mail: weiwei\_song@hnu.edu.cn; shutao\_li@hnu.edu.cn). \par
J. A. Benediktsson is with the Faculty of Electrical and Computer Engineering,
University of Iceland, 101 Reykjavk, Iceland (e-mail: benedikt@hi.is).}
}

\maketitle

\begin{abstract}

%Driven by the urgent demand for analyzing and processing remote sensing big data, large-scale remote sensing image retrieval (RSIR) and classification (RSIC) attracts increasing attention in the remote sensing field. Though RSIR and RSIC are two different tasks, the retrieval and classification performances both highly depend on the representation of feature. In this paper, a unified framework based on deep hashing learning is proposed for RSIR and RSIC for the first time. Specifically, we develop a novel deep hashing convolutional neural network (DHCNN) to learn compact binary features for efficient retrieval and discriminative features for accurate classification for remote sensing images, respectively. In more detailed, a convolutional neural network (CNN) is used to extract high-dimension deep features. Then, a hash layer is perfectly inserted into network to transfer the high-dimension real-value features extracted by CNN into low-dimension binary features, which can significantly improve the speed of image retrieval. In addition, a softmax classifier is used to generate the classification result. In order to make our DHCNN capture more useful information existed in remote sensing images, we elaborately design a loss function to simultaneously consider the semantic information of each image and similarity information of pairs of images. With this learning scheme, the proposed framework achieves the state-of-art performance for RSIR and RSIC on several remote sensing datasets.

Driven by the urgent demand for managing remote sensing big data, large-scale remote sensing image retrieval (RSIR) attracts increasing attention in the remote sensing field. In general, existing retrieval methods can be regarded as visual-based retrieval approaches which search and return a set of similar images from a database to a given query image. Although retrieval methods have achieved great success, there is still a question that needs to be responded to: Can we obtain the accurate semantic labels of the returned similar images to further help analyzing and processing imagery? Inspired by the above question, in this paper, we redefine the image retrieval problem as visual and semantic retrieval of images. Specifically, we propose a novel deep hashing convolutional neural network (DHCNN) to simultaneously retrieve the similar images and classify their semantic labels in a unified framework. In more detail, a convolutional neural network (CNN) is used to extract high-dimensional deep features. Then, a hash layer is perfectly inserted into the network to transfer the deep features into compact hash codes. In addition, a fully connected layer with a softmax function is performed on hash layer to generate class distribution. Finally, a loss function is elaborately designed to simultaneously consider the label loss of each image and similarity loss of pairs of images. Experimental results on two remote sensing datasets demonstrate that the proposed method achieves the state-of-art retrieval and classification performance.

\end{abstract}


\IEEEpeerreviewmaketitle

\section{Introduction}

\IEEEPARstart{W}{ith} the rapid development of remote sensing observation technology, the acquisition of remote sensing images has been largely enhanced not only in volume, but also in resolution. However, these large-scale and high-resolution remote sensing images have also resulted in the significant challenge of how to efficiently manage and analyze the remote sensing big data. Over the past several decades, remote sensing image retrieval (RSIR), which aims to search and return a set of similar images from a database to a given query image, has received increased interest in the remote sensing community.   \par
%is a very important tasks. Though RSIR and RSIC are two different tasks, their performances both highly depend on the power of feature representation. \par

For RSIR, one of the challenges is how to design a retrieval system to return similar images in an accurate and efficient manner. Early retrieval methods for remote sensing images mainly exploited manually annotated tags, e.g., geographical location, acquisition time, or sensor type, to search similar images. This kind of approach, called text-based image retrieval, usually obtains imprecise retrieval results since the visual information of images can not fully represented via annotated tags. In contrast, content-based image retrieval (CBIR) which employs the features extracted directly from the images for retrieval tasks has achieved a great success in recent years \cite{hashing-scalable}. A CBIR system generally consists of two components: (1) feature extraction and (2) a similarity measure. The extracted features for RSIR can be divided into three types: low-level, mid-level, and high-level features. Designing a low-level feature descriptor requires engineering skills and domain expertise. Various low-level features have been exploited in RSIR, such as spectral features \cite{spectral}, texture features \cite{xia-texture}, and shape features \cite{shape1, shape2}. More advanced, mid-level features exhibit superiority over low-level features in representing remote sensing images by exploiting powerful encoding techniques, e.g., bag-of-visual words (BoVW) \cite{BoVW}, Fisher vector (FV) \cite{FV}, and vector of locally aggregated descriptors (VLAD) \cite{VLAD}. However, the above features belong to hand-crafted features which are limited to accurately describe the semantic information that exists in remote sensing images.   \par

Recently, deep learning has made great breakthrough in the computer vision field due to its powerful ability for feature extraction \cite{Alexnet, RCNN, FCN}. Motivated by those successful applications, deep learning has been introduced in the field of remote sensing, including hyperspectral image classification \cite{HSCI_DL, DFFN, SPDF-SVM} and remote sensing scene recognition \cite{MSCP, D-CNN}. In addition, researchers have also attempted to take advantage of high-level features extracted from deep neural networks for RSIR \cite{xia-review}. Specifically, the deep features derived from convolutional neural networks (CNNs) were used to represent remote sensing images and further retrieval relevant images \cite{Low-Dim-CNN, visual-CNN, delving-CNN}. Nevertheless, most existing retrieval methods, including hand-crafted features-based methods and deep features-based methods, adopt Euclidean distance as similarity criteria, which is no longer suitable for real-time retrieval goal due to the time-consuming computation. In order to overcome the above problem, hashing methods have been largely developed for RSIR \cite{hashing-scalable}. Hashing methods aim to learn a set of hash functions to encode the high-dimensional image features into low-dimensional Hamming space, where each image is represented by a binary hash code. By generating a hash-code table for all images, the retrieval can be easily completed via hash lookup or Hamming ranking. More advanced, deep hashing-based methods which take full advantages of deep networks and hashing learning deliver a better performance for RSIR. For examples, Li \emph{et al.} proposed a deep hashing neural network (DHNN) for large-scale RSIR \cite{DHNN}. In such a method, a deep feature learning neural network and a hashing learning neural network were used for high-level semantic feature representation and compact hash code representation, respectively. In \cite{source-invariant-DHCNN}, cross-source remote sensing image retrieval was investigated via source-invariant deep hashing convolutional neural networks (SIDHCNN). In \cite{deep-metric-hash}, a metric and hash-code learning network (MHCLN) was proposed to learn a semantic-based metric space, while simultaneously producing binary hash codes for fast and accurate retrieval of remote sensing images in large archives.  \par

However, existing image retrieval approaches that only return similar images to a given query image from a database may no longer meet the need for further image analyzing and processing. Let us consider this question: Given a query image, whether can we return similar images, at same time, obtain their semantic labels? Actually, solving this problem has a very significant meaning in practical applications. Most importantly, it can further reflect and validate the retrieval performance. In addition, we can better explore the database with semantic labels of returned similar images, which is important for image retrieval in those databases with a few ground-truth samples.   \par

In this paper, we redefine the traditional image retrieval problem as visual and semantic retrieval of images, which aims to retrieve the similar images and simultaneously classify their semantic labels. To this end, we propose a novel deep hashing CNN (DHCNN) to learn compact hash codes for efficient RSIR and discriminative features for accurate semantic label classification. In more detail, we first adopt a CNN to extract high-dimensional deep features from raw remote sensing images. Then, a hash layer is perfectly inserted into the CNN to encode the high-dimensional deep features to low-dimensional hash codes. In addition, a fully connected layer with a softmax function is performed on hash layer to generate class distribution. Finally, we elaborately design a loss function to train DHCNN, where the label information of each image and similarity information of pairs of images are simultaneously considered to improve the ability of representation of features. Once DHCNN is trained enough, for a query image, we can generate its hash code by binarizing the output of hash layer, then, the retrieval can be easily completed via Hamming distance ranking. In addition, the semantic labels of images, including the query image and its similar images, can be obtained by feeding their semantic features into the softmax classifier.   \par

The main contributions of this paper can be summarized as follows.  \par
1) We redefine the image retrieval problem as visual and semantic retrieval of images. To our knowledge, this is the first time to simultaneously retrieve and classify remote sensing images in a unified framework.   \par
2) A novel DHCNN is proposed for fast and efficient RSIR. In such network, a CNN is used to extract deep features and a hash layer is exploited to enforce the continuous-value features to discrete-value hash codes (i.e., -1/+1).  \par
3) Different from existing deep hashing methods that only exploit similar information between samples, we elaborately design an object function which incorporates the label information of each image and similarity information of pairs of images to enhance the representation of features.   \par
%In the new feature space, images from same classes are mapped closely to each other and the images from different classes are mapped far apart.  \par

The rest of this paper is organized as follows. Section II briefly introduces the preliminary knowledge including CNNs and hashing learning. Section III describes the proposed method in detail. The comprehensive experimental results and the corresponding analysis are presented in Section IV. Finally, Section V makes some concluding remarks. \par

\section{Preliminary Knowledge}

\begin{figure*}
\begin{center}
\includegraphics[width=160mm]{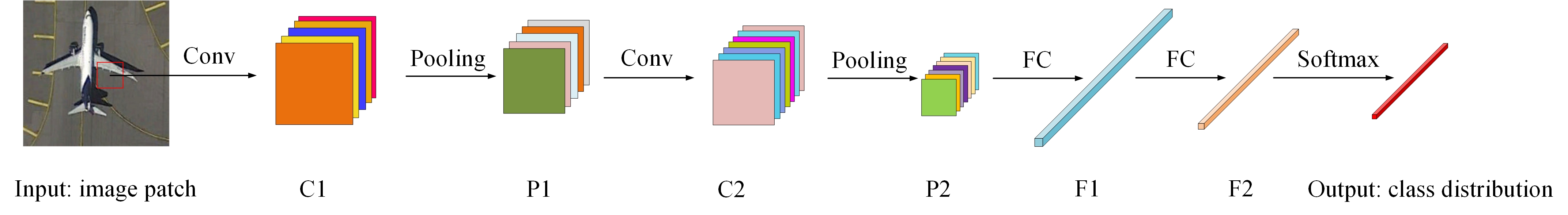}
\end{center}
\caption{The architecture of a typical CNN, which consists of two convolutional-pooling layers, two fully connected layers, and a softmax layer.}
\label{CNN}
\end{figure*}

In this paper, we aim to take advantages of CNNs and hashing learning to enhance the representation of features. In such new feature space, images from same classes are mapped closely to each other and images from different classes are mapped far apart. In this section, we will briefly introduce some preliminary knowledge including CNNs and hashing learning.

\subsection{CNNs}

Recently, CNNs have made great breakthrough in many fields, e.g., image classification \cite{Alexnet}, object detection \cite{RCNN} and semantic segmentation \cite{FCN}. In contrast to fully connected networks, CNNs make use of local connections to extract the features of images. In addition, network parameters can be significantly reduced via the weight-share mechanism. A typical CNN mainly consists of a stack of alternating convolutional layers and pooling layers with a number of fully connected layers. With the rapid development of CNNs, there are many new types of layers, such as dropout layers and local response normalization (LRN) layers \cite{Alexnet}, have been developed. Fig. \ref{CNN} shows a representative structure of CNNs, where the LRN layer and dropout layer are ignored.  \par

Suppose $\mathbf{X}^{l-1}$ be the input of a convolutional layer, the output of this layer can be computed by
\begin{equation}
\mathbf{X}^l = \sum_{i=1}^D \sigma(\mathbf{X}_i^{l-1}*{\mathbf{W}^l}+\mathbf{b}^l)
\end{equation}
where $D$ is the number of channels of $\mathbf{X}^{l-1}$, $\mathbf{W}^l$ and $\mathbf{b}^l$ are the weights and bias, respectively. The operator $*$ represents discrete convolution operation, and $\sigma$ refers to activation function which is utilized to improve nonlinearity of network. Subsequently, a pooling layer may be inserted after convolutional layers to reduce the spatial size
of the feature maps, which can improve the robustness of features. Specifically, for a specific window denoted as $\mathbf{p}$, the averaging pooling operation can be denoted as
\begin{equation}
z=\frac{1}{T} \sum_{(i,j)\in{\mathbf{p}}}x_{ij}
\end{equation}
where $T$ is the number of elements of $\mathbf{p}$, and $x_{ij}$ is the activation value corresponding to the position $(i,j)$. Finally, all features of the previous layer are combined in fully connected layer to extract abstract semantic features. In addition, a softmax function is used on the last fully connected layer to generate the probability distribution of classes. \par

In the past several years, there are many powerful CNNs, e.g., AlexNet \cite{Alexnet}, CaffeNet \cite{CaffeNet}, GoogLeNet \cite{GoogleNet}, VGG \cite{VGG}, and ResNet \cite{DRN}, that have been developed. Although these networks are trained on natural image dataset (i.e., ImageNet \cite{ImageNet}), the extracted features still exhibit powerful generalization ability on remote sensing datasets \cite{Do_deep_features}. In addition, considering that the available training samples are relatively small and labeling unknown samples is very difficult, we transfer existing deep CNNs to our DHCNN to reduce the need on training samples.   \par

\subsection{Hashing Learning}

Due to its encouraging efficiency in both speed and storage, hashing technique has been widely used in large-scale image retrieval \cite{Hierarchical-RNN-Hashing, Laten-SMH, Supervised-DH}. Given a training set of $N$ points $\left\{\mathbf{x}_i\right\}_{i=1}^N$, each point is represented as $D$-dimensional feature vector. The goal of hashing is to learn nonlinear function $f: \mathbf{x}\longmapsto h\in\left\{-1, 1\right\}^K $ to encode each point $\mathbf{x}$ in compact $K$-bit hash code $h = f(\mathbf{x})$. Existing learning-based hashing methods can be roughly divided into two categories: unsupervised hashing and supervised hashing.  \par

Unsupervised hashing methods use the unlabeled training data to learn a set of hash functions that can encode input data points to binary codes. The representative methods include spectral hashing (SH) \cite{SH}, iterative quantization (ITQ) \cite{ITQ}, and discrete graph hashing (DGH) \cite{DGH}. In contrast, supervised hashing aims to generate similarity-preserving representations with shorter hash codes by utilizing supervised information, e.g., point-wise labels, pairwise labels, and ranking labels. In past several years, there are many successful supervised hashing methods that have been developed for fast image retrieval, including binary reconstruction embedding (BRE) \cite{BRE}, minimal loss hashing (MLH) \cite{MLH}, sparse embedding and least variance encoding (SELVE) \cite{SELVE}, and supervised hashing with kernels (KSH) \cite{KSH}. By utilizing the supervised information, images from same classes have small feature distance while images from different classes have large features distance in Hamming space. \par

Most of the existing hashing methods use hand-crafted visual features to encode each input image, which may degrade their hashing performance because hand-crafted features do not necessarily capture accurate similarity of images. Recently, many researches have focused on integrating the hashing technique into CNNs, which delivers satisfying performance for image retrieval. For examples, Xia \emph{et al.} proposed a two-stage method to train a CNN to fit binary
codes computed from the pairwise similarity matrix \cite{CNNH}. Li \emph{et al.} performed simultaneous feature learning and hash code learning for image retrieval with pairwise labels \cite{DPSH}. In \cite{Bit-scalable}, a deep hashing with regularized similarity learning framework was proposed to generate compact and bit-scalable hashing codes for image retrieval and person re-identification. In addition, Zhang \emph{et al.} focused on the problem of unsupervised deep hashing and discovered pseudo labels to train a deep network for scalable image retrieval \cite{Unsupervised-pseudo-labels}. \par

\section{Proposed Framework}

\begin{figure*}
\begin{center}
\includegraphics[width=160mm]{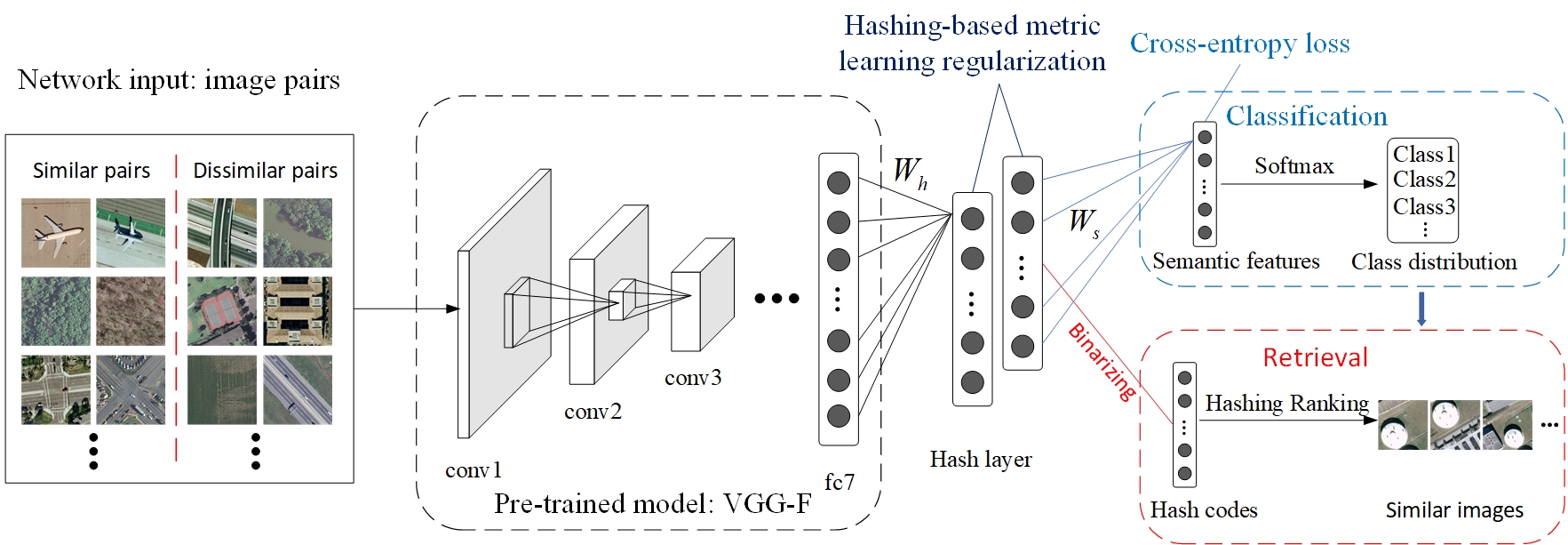}
\end{center}
\caption{The proposed DHCNN for visual and semantic retrieval of remote sensing images. Firstly, a pre-trained CNN is introduced to extract deep features. Then, a hash layer with metric learning regularization is used to transfer high-dimensional deep features into low-dimensional hash codes. Furthermore, a fully connected layer with a softmax classifier is used to generate class distribution. With the hash codes and class distribution, a set of similar images to the given query image and their semantic labels are easily obtained via Hashing ranking.}
\label{DHCNN}
\end{figure*}

%In this section, we propose a unified framework based on DHCNN simultaneously for RSIR and RSIC.
In order to cope with the challenge of high intraclass and low interclass variabilities that exist in remote sensing images, we combine the deep leaning and hashing learning to minimize the feature distance between similar image pairs and maximize the feature distance between dissimilar image pairs. To this end, we design an object function to simultaneously consider the label information of each image and similarity function of pairs of images. Through the proposed DHCNN, we can extract discriminative semantic features for accurate classification and learn compact hash codes for efficient retrieval. Fig. \ref{DHCNN} illustrates the proposed DHCNN which consists of a pre-trained CNN, a hash layer, and a fully connected layer with a softmax classifier. In the following part, we will introduce the proposed method in detail.  \par

\subsection{Deep Feature Extraction}

\begin{table}
\centering
\footnotesize
  \caption{The Configuration of Deep Network Used in Deep Feature Extraction, Which Is Transferred From VGG-F \cite{CNN-S}}
  \label{CNN-S}
\begin{tabular}{cc}
\hline
\hline
 Layer          & Configuration    \\
\hline
conv1        & filter $96\times{7}\times{7}$, stride $2\times{2}$, pad 0, LRN, pool $3\times{3}$  \\
conv2        & filter $256\times{5}\times{5}$, stride $1\times{1}$, pad 1, pool $2\times{2}$ \\
conv3        & filter $512\times{3}\times{3}$, stride $1\times{1}$, pad 1 \\
conv4        & filter $512\times{3}\times{3}$, stride $1\times{1}$, pad 1 \\
conv5        & filter $512\times{3}\times{3}$, stride $1\times{1}$, pad 1, pool $3\times{3}$ \\
full6        & 4096, dropout \\
full7        & 4096, dropout \\
\hline
\hline
\end{tabular}
\end{table}

In general, training a CNN from scratch requires a large number of training samples to learn the model parameters. However, in the remote sensing field, the available training samples are relatively small and labeling unknown samples is a costly and time-consuming work. To solve this problem, we adopt a pre-trained CNN model to decrease the burden on training samples. Here, we take VGG-F \cite{CNN-S} model as example to explicate the part of deep feature extraction, which is illustrated in the black dotted box in Fig. \ref{DHCNN}. Specifically, the network parameters of VGG-F \cite{CNN-S}, including the first five convolutional layers and two fully connected layers, are transferred to our DHCNN. The detailed configuration is shown in Table \ref{CNN-S}. For convolutional layers, ``filter'' specifies the number of convolution filters and kernel size; ``stride'' and ``pad'' indicate the convolutional strides and spatial padding, respectively; ``LRN'' refers to Local Response Normalization \cite{Alexnet}; ``pool'' specifies the max-pooling size. For fully connected layers, ``4096'' indicates the feature dimension and the dropout technique \cite{Alexnet} is applied to full6 and full7. The activation function for all weight layers is the REctification Linear Unit (ReLU) \cite{Alexnet}.  \par

Supposed there are $N$ training samples denoted as $\mathbf{X}=\left\{\mathbf{x}_i\right\}_{i=1}^N$, the corresponding set of labels can be represented as $\mathbf{Y}=\left\{\mathbf{y}_i\right\}_{i=1}^N$, where $\mathbf{y}_i\in\mathbb{R}^C$ is the ground-truth vector of sample $\mathbf{x}_i$ with only one element being 1 and others being 0, $C$ is the total number of image scene classes. For arbitrary remote sensing image $\mathbf{x}_i\in\mathbf{X}$, we can extract its deep features (i.e., the output of the fc7 layer) denoted as $\mathbf{f}_i$ by

\begin{equation}
\mathbf{f}_i=\Phi(\mathbf{x}_i; \theta), i=1,2,...N
\end{equation}
where $\Phi$ is the network function characterized by the $\theta$ which denotes all the parameters of the first seven layers of VGG-F \cite{CNN-S}. This propagation actually performs a series of nonlinear and linear transformations, including convolution, pooling, and nonlinear mapping.  \par

%Supposed there are $L$ layers in DHCNN, the output of each layer can be represented by
%\begin{equation}
%\begin{aligned}
%&\mathbf{O}^L=\sigma^L(\mathbf{W}^L\mathbf{O}^{L}+\mathbf{b}^L)\\
%&\mathbf{O}^l=\sigma^l(\mathbf{W}^l\mathbf{O}^{l}+\mathbf{b}^l), l=1,2,...L-1
%\end{aligned}
%\end{equation}
%where $\sigma^L$ is the softmax function, $\sigma^l$ is the ReLU function, $\mathbf{W}=\left\{\mathbf{W}^1, \mathbf{W}^2, ..., \mathbf{W}^L\right\}$ and $\mathbf{b}=\left\{\mathbf{b}^1, \mathbf{b}^2, ..., \mathbf{b}^L\right\}$ are the filter weights and biases of network, respectively. Here, we use $\textit{f}$ to represent the deep features extracted from $L-2$th layer, i.e., $O^L-2$, for more clear description. \par

%After obtaining the class distribution $\mathbf{O}^L$, we adopt cross-entropy loss to minimize the error between the predicted label and ground-truth label. This loss function is defined by as

%\begin{equation}
%J_1(\mathbf{X}, \mathbf{W}, \mathbf{b})=-\frac{1}{N}\sum_{i=1}^{N}<\mathbf{y}_i, \mathbf{O}^L(\mathbf{x}_i)>
%\end{equation}
%where $<>$ represents inner production operation. By minimizing the object function $J_1$, the CNN can learn discriminative semantic features of each images.    \par

\subsection{Hashing-based Metric Learning}

In light of the large intraclass and low interclass variabilities that exist in remote sensing images, we adopt hashing-based metric learning to constrain images from same classes to be encoded as closely as possible and images from different classes to be encoded far away each other in feature space. To this end, we use pairwise input to train our network, which can explore the similarity/disimilarity information between images. Let $(\mathbf{x}_i,\mathbf{x}_j)$ be a pair of images, we define its label $s_{ij}$ such that $s_{ij}=1$ if $\mathbf{x}_i$ and $\mathbf{x}_j$ come from same class and $0$ otherwise. As mentioned above, we can easily obtain their deep features $(\mathbf{f}_i, \mathbf{f}_j)$ via forward propagation. Subsequently, a hash layer is inserted after the pre-trained CNN to transfer the high-dimensional deep features into compact $K$-bit hash codes, which can be formulated as
\begin{equation}
\mathbf{b}_t=sgn(\mathbf{u}_t), t=i,j
\end{equation}
where $\mathbf{u}_t=\mathbf{W}_h\mathbf{f}_t+\mathbf{v}_h$ is the hash-like feature, $\mathbf{W}_h\in\mathbb{R}^{K\times{4096}}$ denotes a weight matrix, $\mathbf{v}_h\in\mathbb{R}^{K\times{1}}$ denotes a bias vector, $sgn(\cdot)$ performs element-wise operations for a matrix or a vector, i.e., $sgn(x)=1$ if $x>0$ and $-1$ otherwise.  \par

Once obtaining the hash codes $\mathbf{B}=\left\{{\mathbf{b}_t}\right\}_{t=1}^N$ for all the samples, the likelihood of the pairwise labels ${S}=\left\{s_{ij}\right\}$ can be defined as

\begin{equation}
p(s_{ij}|\mathbf{B})=
\left\{
\begin{array}{lr}
\varphi(\omega_{ij}),       & s_{ij}=1 \\
1-\varphi(\omega_{ij})£¬    & s_{ij}=0
\end{array}
\right.
\end{equation}
where $\varphi(\cdot)$ is the logistic function and $\varphi(x)=\frac{1}{1+e^{-x}}$, $\omega_{ij}=\frac{1}{2}\mathbf{b}_i^T\mathbf{b}_j$. Based on the above definition, the loss function can be given by taking the negative log-likelihood of the observed pairwise labels in $S$
\begin{equation}
\begin{split}
\mathcal{L}_1 & = -log\ p({S}|\mathbf{B})=-\sum_{s_{ij}\in{S}} log\ p(s_{ij}|\mathbf{B}) \\
            & = -\sum_{s_{ij}\in{S}} (s_{ij}\omega_{ij}-log(1+e^{\omega_{ij}})) .   \\
\end{split}
\label{L1}
\end{equation}
However, directly solving the problem (\ref{L1}) is very hard due to the discrete values in formulation. Motivated by \cite{DPSH}, the above loss function can be reformulated in a discrete way
\begin{equation}
\begin{split}
\mathcal{L}_2 & = -\sum_{s_{ij}\in{S}} (s_{ij}\psi_{ij}-log(1+e^{\psi_{ij}})) \\
            & +\beta \sum_{i=1}^N \left\|\mathbf{u}_i-\mathbf{b}_i\right\|_2^2    \\
\end{split}
\label{L2}
\end{equation}
where $\psi_{ij}=\frac{1}{2}\mathbf{u}_i^T\mathbf{u}_j, i,j=1,2,...,N$, $\beta$ is a regularization parameter which can constrain $\mathbf{u}_i$ approach to $\mathbf{b}_i$. Through minimizing the $\mathcal{L}_2$, the feature distance in Hamming space (i.e., Hamming distance) between similar samples can be optimized to be as small as possible, and the Hamming distance between dissimilar samples becomes as large as possible.  \par

\subsection{Object Function and Solving}

Different from existing deep hashing methods for image retrieval that only utilize similarity information between images to learn hash codes \cite{DPSH, CNNH, DHNN, source-invariant-DHCNN, deep-metric-hash}, we also consider semantic information of each image to further improve the ability of feature representation. To this end, a fully connected layer with a softmax function is added after the hash layer to generate the class distribution for each image. This procedure can be represented by
\begin{equation}
\mathbf{t}_k=softmax(\mathbf{W}_s\mathbf{u}_k+\mathbf{v}_s), k=1,2,...,N
\end{equation}
where $\mathbf{W}_s\in\mathbb{R}^{C\times{K}}$  and $\mathbf{v}_s\in\mathbb{R}^{C\times{1}}$ denote the weight matrix and bias vector, respectively. Then, we adopt cross-entropy loss to minimize the error between the predicted label and ground-truth label
\begin{equation}
\mathcal{L}_3=-\frac{1}{N}\sum_{i=1}^{N}<\mathbf{y}_i, log\mathbf{t}_i>
\label{L3}
\end{equation}
where $<>$ represents inner production operation. By minimizing the loss function $\mathcal{L}_3$, the CNN can learn discriminative semantic features of each images.  \par

As mentioned above, loss function $\mathcal{L}_2$ aims to learn the similarity information between images, $\mathcal{L}_3$ aims to learn the label information of each image. Here, we design a new loss function to simultaneously consider the similarity information and label information to improve the network performance. This new loss function is defined as
\begin{equation}
\mathcal{L}_4=\eta\mathcal{L}_2+(1-\eta)\mathcal{L}_3
\label{L4}
\end{equation}
where $\eta\in[0,1]$ is the regularization parameter to balance the label information and similarity information. Specifically, when $\eta=0$, the object function only utilizes label information of each images. On the other hand, only the similarity information between images is considered when $\eta=1$. Finally, our object function is to minimize the loss function $\mathcal{L}_4$, i.e.,
\begin{equation}
\begin{aligned}
\mathcal{J}= \text{min}\mathcal{L}_4= & \text{min}\left\{\eta(-\sum_{s_{ij}\in{S}}
  (s_{ij}\psi_{ij}-log(1+e^{\psi_{ij}}))\right. \\
     & +\beta \sum_{i=1}^N \left\|\mathbf{u}_i-\mathbf{b}_i\right\|_2^2) \\
     & \left. +(1-\eta)(-\frac{1}{N}\sum_{i=1}^{N}<\mathbf{y}_i, log\mathbf{t}_i>) \right\} .
\end{aligned}
\label{J}
\end{equation}

In order to solve the problem in (\ref{J}), we adopt the stochastic gradient descent (SGD) algorithm to learn the parameters, including $\mathbf{W}_h$, $\mathbf{W}_s$, $\mathbf{v}_h$, $\mathbf{v}_s$, and $\theta$. At first. we compute the gradients of the objective function $\mathcal{J}$ with respect to $\mathbf{t}_i$ and $\mathbf{u}_i$, which can be represented by

\begin{equation}
\frac{\partial{\mathcal{J}}}{\partial{\mathbf{t}_i}}=(1-\eta)\frac{\partial{\mathcal{L}_3}}
{\partial{\mathbf{t}_i}}=-(1-\eta)\frac{1}{N}\frac{1}{\mathbf{t}_i}
\end{equation}

\begin{equation}
\begin{split}
 \frac{\partial{\mathcal{J}}}{\partial{\mathbf{u}_i}} & = \eta\frac{\partial{\mathcal{L}_2}}
{\partial{\mathbf{u}_i}}+(1-\eta)\frac{\partial{\mathcal{L}_3}}{\partial{\mathbf{u}_i}}= \eta(\frac{1}{2}\sum_{j:s_{ij}\in{S}}(a_{ij}-s_{ij})\mathbf{u}_j  \\
&+\frac{1}{2}\sum_{j:s_{ji}\in{S}}(a_{ji}-s_{ji})\mathbf{u}_j
+2\beta(\mathbf{u}_i-\mathbf{b}_i)) \\
& + (1-\eta)(-\frac{1}{N}\mathbf{W}_s^T(\mathbf{y}_i-\mathbf{t}_i))
\end{split}
\end{equation}
where $a_{ij}=\varphi(\frac{1}{2}\mathbf{u}_i^T\mathbf{u}_j)$. Then we can further compute gradients of the objective function $\mathcal{J}$ with respect to $\mathbf{W}_s$, $\mathbf{v}_s$, $\mathbf{W}_h$, $\mathbf{v}_h$, and $\theta$ by utilizing chain rule

\begin{equation}
 \frac{\partial{\mathcal{J}}}{\partial{\mathbf{W}_s}}=\frac{\partial{\mathcal{J}}}{\partial{\mathbf{t}_i}}
 \frac{\partial{\mathbf{t}_i}}{\partial{\mathbf{o}_i}}\frac{\partial{\mathbf{o}_i}}{\partial{\mathbf{W}_s}}
 =\frac{\partial{\mathcal{J}}}{\partial{\mathbf{t}_i}}\odot\mathbf{t}_i\odot(\mathbf{y}_i-\mathbf{t}_i)\mathbf{u}_i^T
\end{equation}

\begin{equation}
\frac{\partial{\mathcal{J}}}{\partial{\mathbf{v}_s}}=\frac{\partial{\mathcal{J}}}{\partial{\mathbf{t}_i}}
 \frac{\partial{\mathbf{t}_i}}{\partial{\mathbf{o}_i}}\frac{\partial{\mathbf{o}_i}}{\partial{\mathbf{v}_s}}
 =\frac{\partial{\mathcal{J}}}{\partial{\mathbf{t}_i}}\odot\mathbf{t}_i\odot(\mathbf{y}_i-\mathbf{t}_i)
\end{equation}

\begin{equation}
\frac{\partial{\mathcal{J}}}{\partial{\mathbf{W}_h}}=\frac{\partial{\mathcal{J}}}{\partial{\mathbf{u}_i}}
 \frac{\partial{\mathbf{u}_i}}{\partial{\mathbf{W}_h}}
 =\frac{\partial{\mathcal{J}}}{\partial{\mathbf{u}_i}}\mathbf{f}_i^T
\end{equation}

\begin{equation}
\frac{\partial{\mathcal{J}}}{\partial{\mathbf{v}_h}}=\frac{\partial{\mathcal{J}}}{\partial{\mathbf{u}_i}}
 \frac{\partial{\mathbf{u}_i}}{\partial{\mathbf{v}_h}}
 =\frac{\partial{\mathcal{J}}}{\partial{\mathbf{u}_i}}
\end{equation}

\begin{equation}
\frac{\partial{\mathcal{J}}}{\partial{\Phi(\mathbf{x}_i;\theta)}}=
\frac{\partial{\mathcal{J}}}{\partial{\mathbf{u}_i}}
 \frac{\partial{\mathbf{u}_i}}{\partial{\Phi(\mathbf{x}_i; \theta)}}=\mathbf{W}_h^T\frac{\partial{\mathcal{J}}}{\partial{\mathbf{u}_i}}
\end{equation}
where $\mathbf{o}_i=\mathbf{W}_s\mathbf{u}_i+\mathbf{v}_s$, the operation $\odot$ denotes element-wise multiplication. Finally, we can update all parameters by using the gradient descent method as follows
\begin{equation}
\xi = \xi - \mu\frac{\partial{\mathcal{J}}}{\partial{\xi}}, \xi=\mathbf{W}_s, \mathbf{W}_h, \mathbf{v}_s, \mathbf{v}_h, \Phi(\mathbf{x}_i; \theta)
\end{equation}
where $\mu$ is learning rate.  \par

\subsection{Retrieval and Classification}

Once the DHCNN is trained enough, we can obtain the hash codes and class labels for all samples from the database. Specifically, for an arbitrary image $\mathbf{x}_q$, its hash code $\mathbf{b}_q$ and class label $c_q$ can be determined by
\begin{equation}
\mathbf{b}_q=sgn(\mathbf{u}_q)=sgn(\mathbf{W}_h\mathbf{f}_q+\mathbf{v}_h)
\end{equation}

\begin{equation}
c_q=\underset{k=1,2,...,C}{\arg\,\max}{\mathbf{t}_i^k}
\end{equation}
where $\mathbf{t}_i^k$ is the $k$th component of vector $\mathbf{t}_i$. Finally, given a query image, a set of similar images and their class labels can be easily returned via ranking Hamming distance between the query image and images from the database.  \par

\section{Experimental Results and Analysis}

To validate the effectiveness of the proposed approach for RSIR and classification, a comprehensive set of experiments are conducted on two remote sensing datasets.  \par
%Our implementation is publicly available at https://github.com/weiweisong415.  \par

\subsection{Datasets and Experimental Settings}

\begin{figure*}
\begin{center}
\includegraphics[width=162mm]{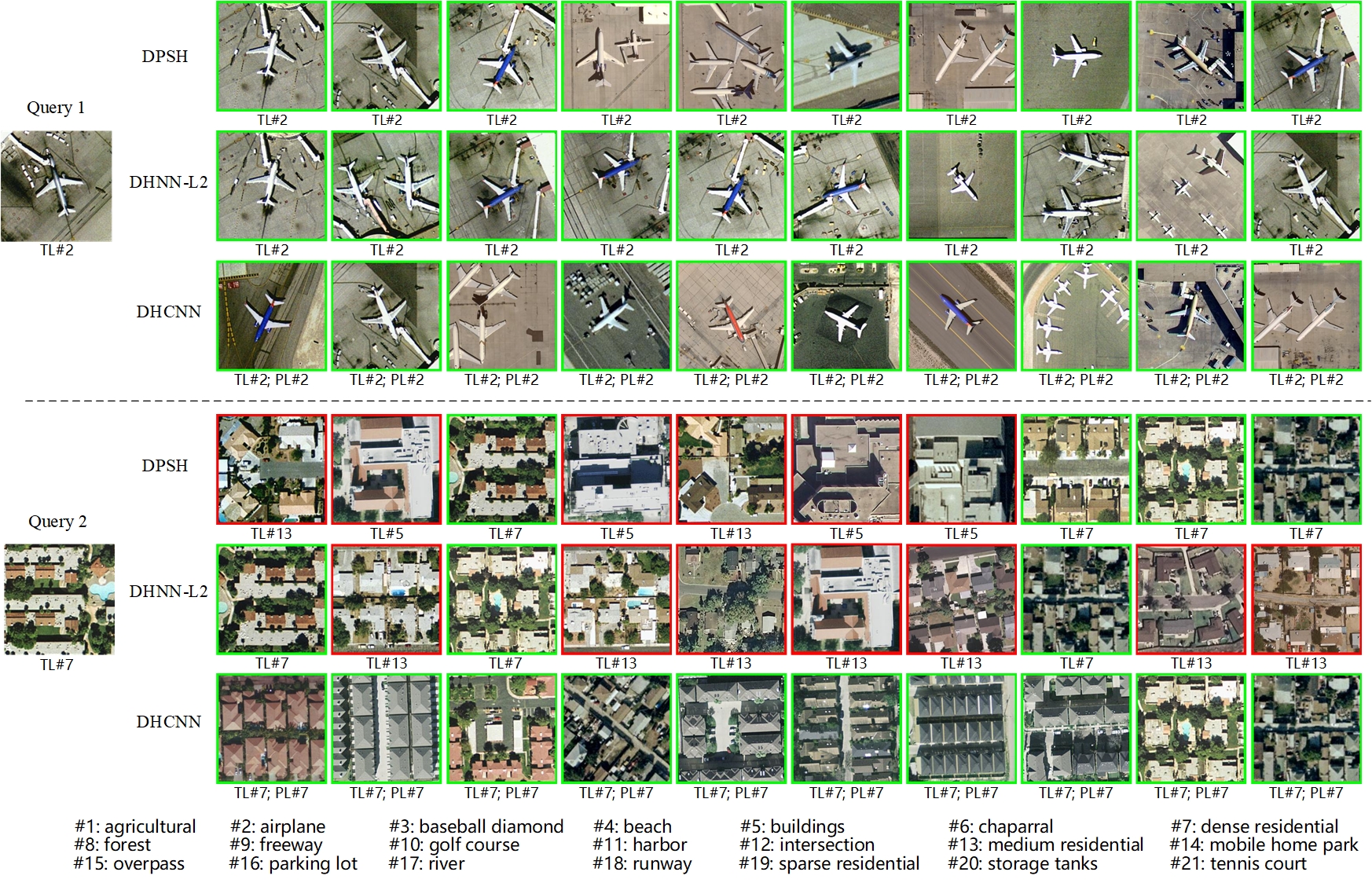}
\end{center}
\caption{Two query examples with top ten retrieved images on the UCMD. For each query example, the top, middle, and bottom rows represent retrieval results obtained by DPSH \cite{DPSH}, DHNN-L2 \cite{DHNN}, and our proposed method DHCNN. The green rectangle marks true positives, while the red rectangle marks false positives. The ``TL" and ``PL" represent the true label and predicted label of images, respectively.}
\label{UCM21-query}
\end{figure*}

\begin{figure*}
\begin{center}
\includegraphics[width=162mm]{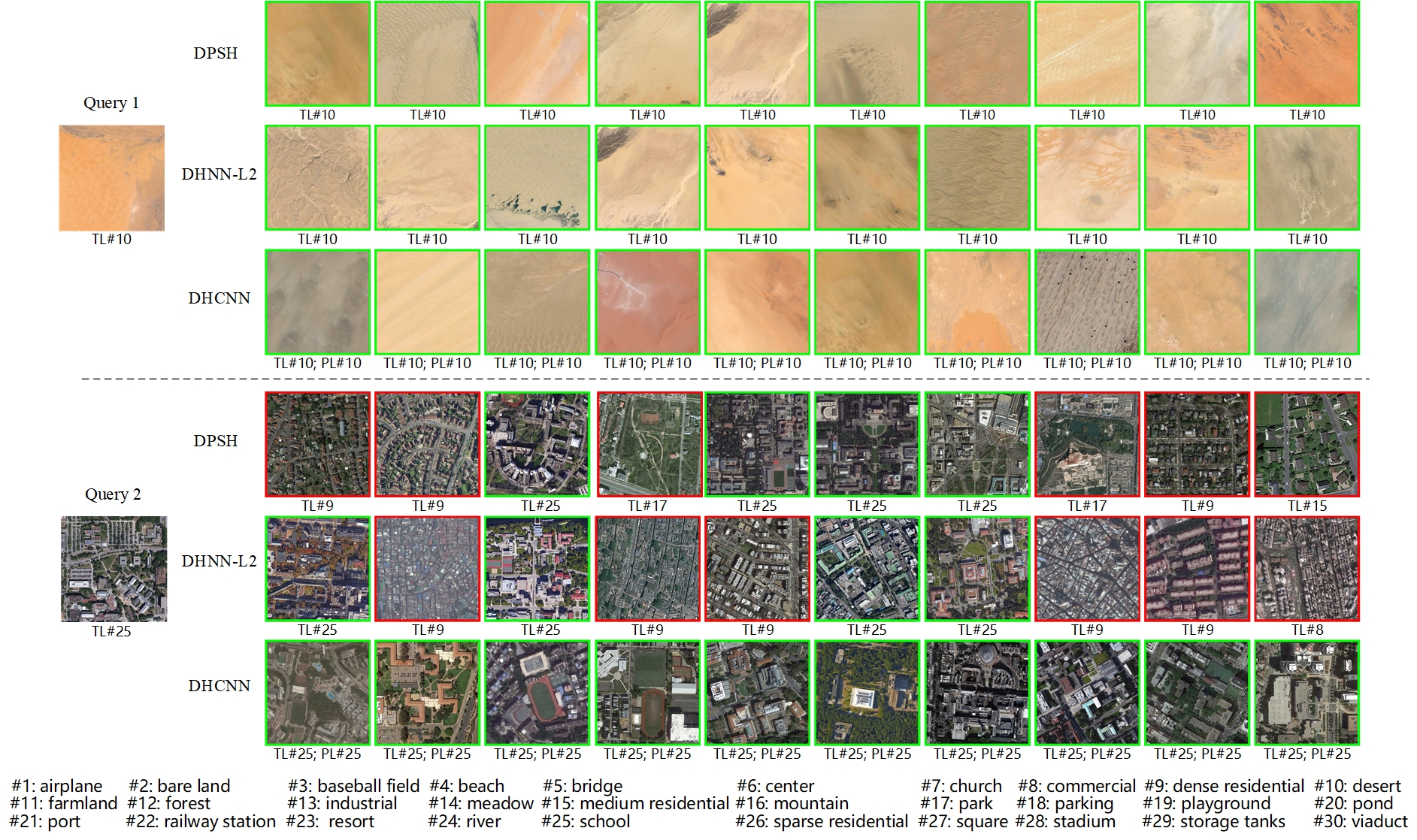}
\end{center}
\caption{Two query examples with top ten retrieved images on the AID. For each query example, the top, middle, and bottom rows represent retrieval results obtained by DPSH \cite{DPSH}, DHNN-L2 \cite{DHNN}, and our proposed method DHCNN. The green rectangle marks true positives, while the red rectangle marks false positives. The ``TL" and ``PL" represent the true label and predicted label of images, respectively.}
\label{AID30-query}
\end{figure*}

1) The University of California, Merced dataset (UCMD) \cite{UCM21} is manually extracted from large images downloaded from the United States Geological Survey (USGS). The UCMD is widely used for evaluating the performance of retrieval and classification for remote sensing images. It contains 21 land cover categories, each category includes 100 images of $256\times{256}$ pixels, and the spatial resolution of each pixel is 0.3 m. Some classes in UCMD are highly overlapping, e.g., medium residential and dense residential, which makes it a challenging
dataset. We randomly select 80 images per class (1680 images in total) as training set to learn hash function. The rest of samples are used to evaluate the retrieval performance.  \par

2) The aerial image dataset (AID) \cite{AID} is a large-scale aerial image dataset which was collected with the goal of advancing the state-of-the-art in scene classification of remote sensing
images. The dataset has a number of 10000 images within 30 classes. Each class consists of 220 to 420 images of size of $600\times{600}$ pixels. The images in AID are from different remote imaging sensors and the spatial resolution varies greatly between around 0.5 to 8 m, which brings more challenges for scene classification and image retrieval than the single source images like UCMD.   \par

%3) The PatternNet \cite{PatternNet} is a large-scale high-resolution remote sensing dataset collected from Google Earth imagery or via the Google Map API for US cities. This dataset has a number of 30400 images within 38 classes. Each class has 800 images with size of $256\times{256}$ pixels. The spatial resolution of images in PatternNet is relatively high (i.e., 0.062 to 4.693 m) so that the classes of interest constitute a larger portion of the images. Different from recent remote sensing datasets which are mainly for classification application, PatternNet is the first large-scale dataset for retrieval application for remote sensing images, especially for deep learning which requires large amounts of labeled data.   \par

We systematically compare our method with some state-of-the-art hashing methods, including SH \cite{SH}, ITQ \cite{ITQ}, SELVE \cite{SELVE}, KSH \cite{KSH}, DPSH \cite{DPSH}, and DHNN with L2 regularization (DHNN-L2) \cite{DHNN}. To make our experiments more convincing, the SH \cite{SH}, ITQ \cite{ITQ}, SELVE \cite{SELVE}, and KSH \cite{KSH} methods are designed both on hand-crafted features and deep features. Specifically, each remote sensing image is represented by 512-dimensional GIST descriptors \cite{GIST} and 4096-dimensional CNN features extracted from fc7 layer of VGG-F \cite{CNN-S}, respectively, for the above methods. For the deep hashing methods, including DPSH \cite{DPSH}, DHNN-L2 \cite{DHNN}, and our proposed DHCNN, we first resize all images to be $224\times{224}$ pixels and then directly feed the raw image pixels into deep networks. It is worth mentioning that we adopt same network architecture (i.e., VGG-F \cite{CNN-S}) for DPSH, DHNN-L2, and our proposed DHCNN to achieve fair comparison. The hyper-parameters $\eta$ and $\beta$ existed in formulation (\ref{J}) are set to be 0.2 and 25 for all datasets, respectively.   \par

\begin{table*}
\small
\centering
%\footnotesize
  \caption{Image Retrieval Results in Terms of MAP (Shown in Percentages) With 16, 32, 48, and 64 Bits on the UCMD and AID. The Scale of Test Query Sets Are 420 (20\% Samples Per Class) and 5000 (50\% Samples Per Class) for UCMD and AID, Respectively. The MAPs Are Computed Using All the Training Sets and the Best Values Are Shown in Boldface. }
%\begin{tabular}{ccp{0.8cm}<{\centering}p{0.8cm}<{\centering}p{0.8cm}<{\centering}p{0.8cm}<{\centering}p{1cm}<{\centering}p{0.8cm}<{\centering}p{0.8cm}<{\centering}p{0.8cm}<{\centering}}
\begin{tabular}{c|cccc|cccc}
\hline
\multirow{2}{*}{Method} & \multicolumn{4}{c|}{UCMD}   & \multicolumn{4}{c}{AID}  \\
\cline{2-9}
   & 16 bits  & 32 bits & 48 bits & 64 bits        & 16 bits & 32 bits & 48 bits  & 64 bits  \\
\hline
\textbf{DHCNN} (our method) & \textbf{96.52} & \textbf{96.98} & \textbf{97.46} & \textbf{98.02} & \textbf{89.05} & \textbf{92.97} & \textbf{94.21} & \textbf{94.27} \\
DHNN-L2 \cite{DHNN} & 67.73 & 78.23 & 82.43 & 85.59 & 57.87 & 70.36 & 73.98 & 77.20 \\
DPSH \cite{DPSH} & 53.64 & 59.33 & 62.17 & 65.21 & 28.92 & 35.30 & 37.84 & 40.78 \\
KSH-CNN \cite{KSH} & 75.50 & 83.62 & 86.55 & 87.22 & 48.26 & 58.15 & 61.59 & 63.26 \\
ITQ-CNN \cite{ITQ} & 42.65 & 45.63 & 47.21 & 47.64 & 23.35 & 27.31 & 28.79 & 29.99 \\
SELVE-CNN \cite{SELVE} & 36.12 & 40.36 & 40.38 & 38.58 & 34.58 & 37.87 & 39.09 & 36.81 \\
DSH-CNN \cite{DSH} & 28.82 & 33.07 & 33.15 & 34.59 & 16.05 & 18.08 & 19.36 & 19.72 \\
SH-CNN \cite{SH} & 29.52 & 30.08 & 30.37 & 29.31 & 12.69 & 16.99 & 16.16 & 16.21 \\
KSH-GIST \cite{KSH} & 38.75 & 43.27 & 44.44 & 46.38 & 18.96 & 22.64 & 24.76 & 26.32 \\
ITQ-GIST \cite{ITQ} & 19.85 & 20.44 & 20.67 & 20.96 & 9.99 & 10.63 & 11.04 & 11.30 \\
SELVE-GIST \cite{SELVE} & 17.24 & 18.18 & 18.08 & 18.17 & 10.37 & 10.62 & 10.62 & 10.40 \\
DSH-GIST \cite{DSH} & 16.57 & 17.79 & 19.21 & 19.18 & 9.35 & 9.53 & 9.74 & 10.05 \\
SH-GIST \cite{SH} & 13.76 & 14.46 & 14.24 & 14.17 & 6.71 & 7.02 & 6.97 & 6.95  \\
\hline
\end{tabular}
\label{Results-MAP}
\end{table*}

\subsection{Evaluation Metrics}

\begin{figure*}
\begin{center}
\subfigure[]{\includegraphics[width=57mm]{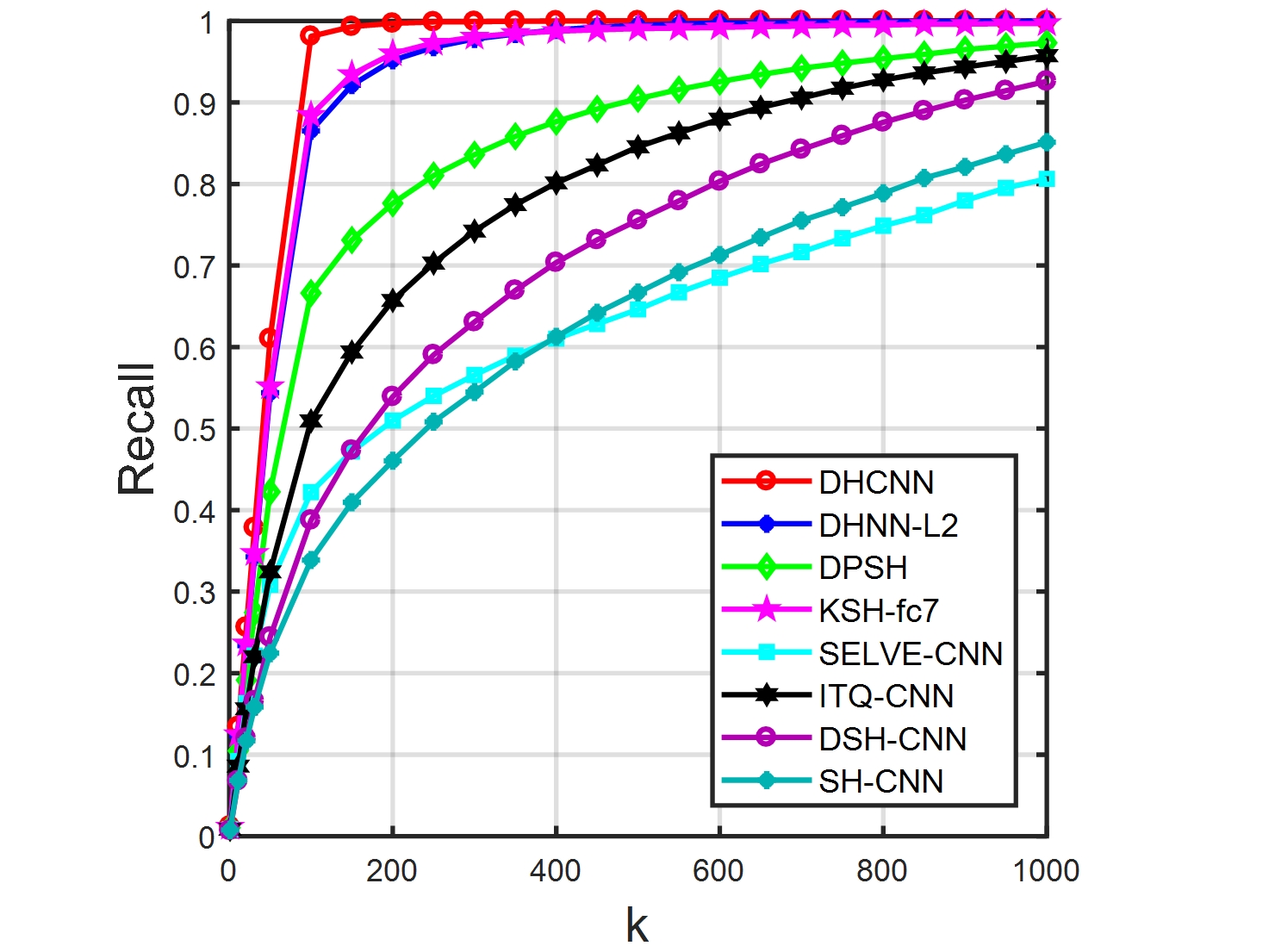}}
\subfigure[]{\includegraphics[width=57mm]{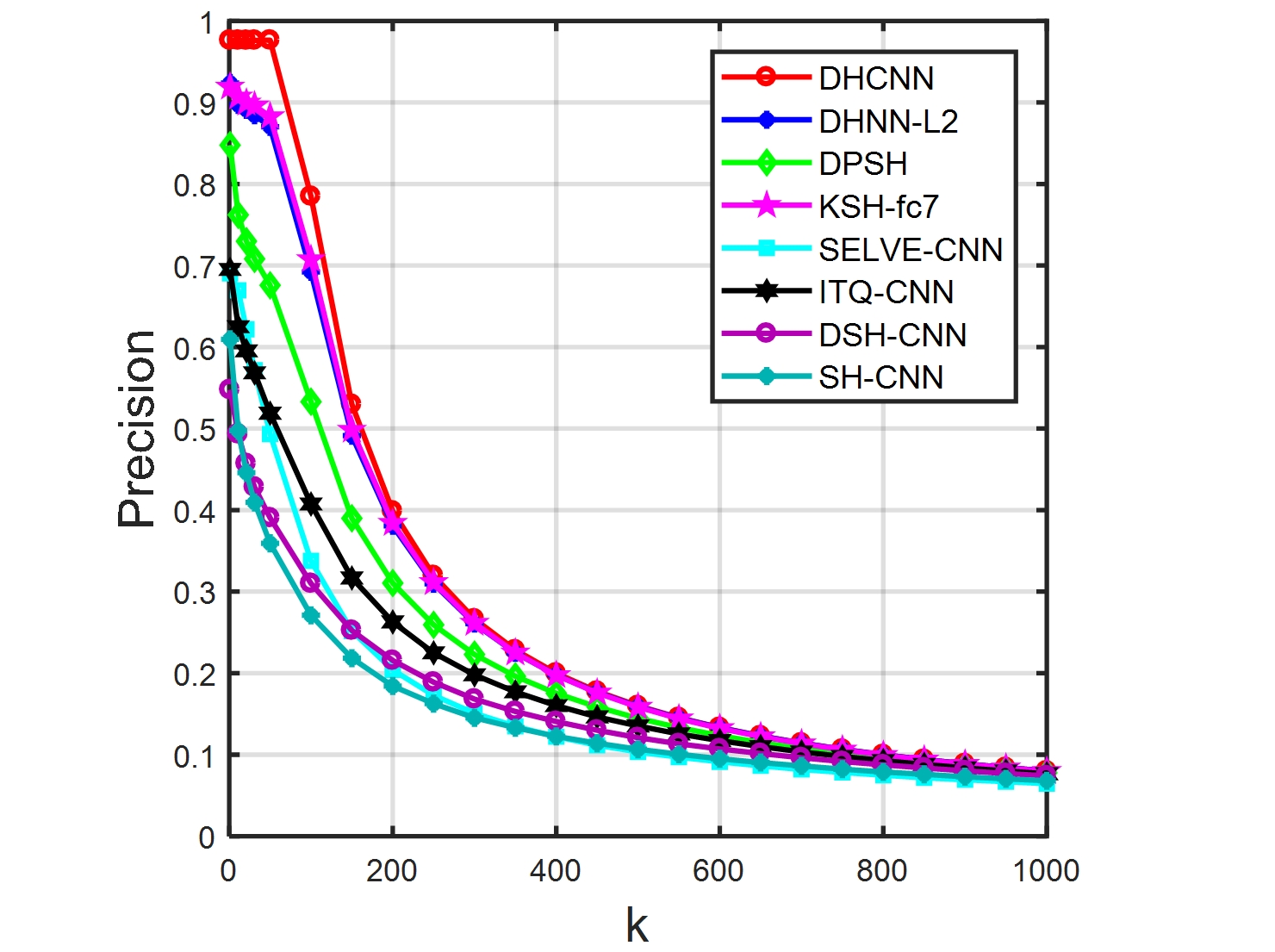}}
\subfigure[]{\includegraphics[width=55mm]{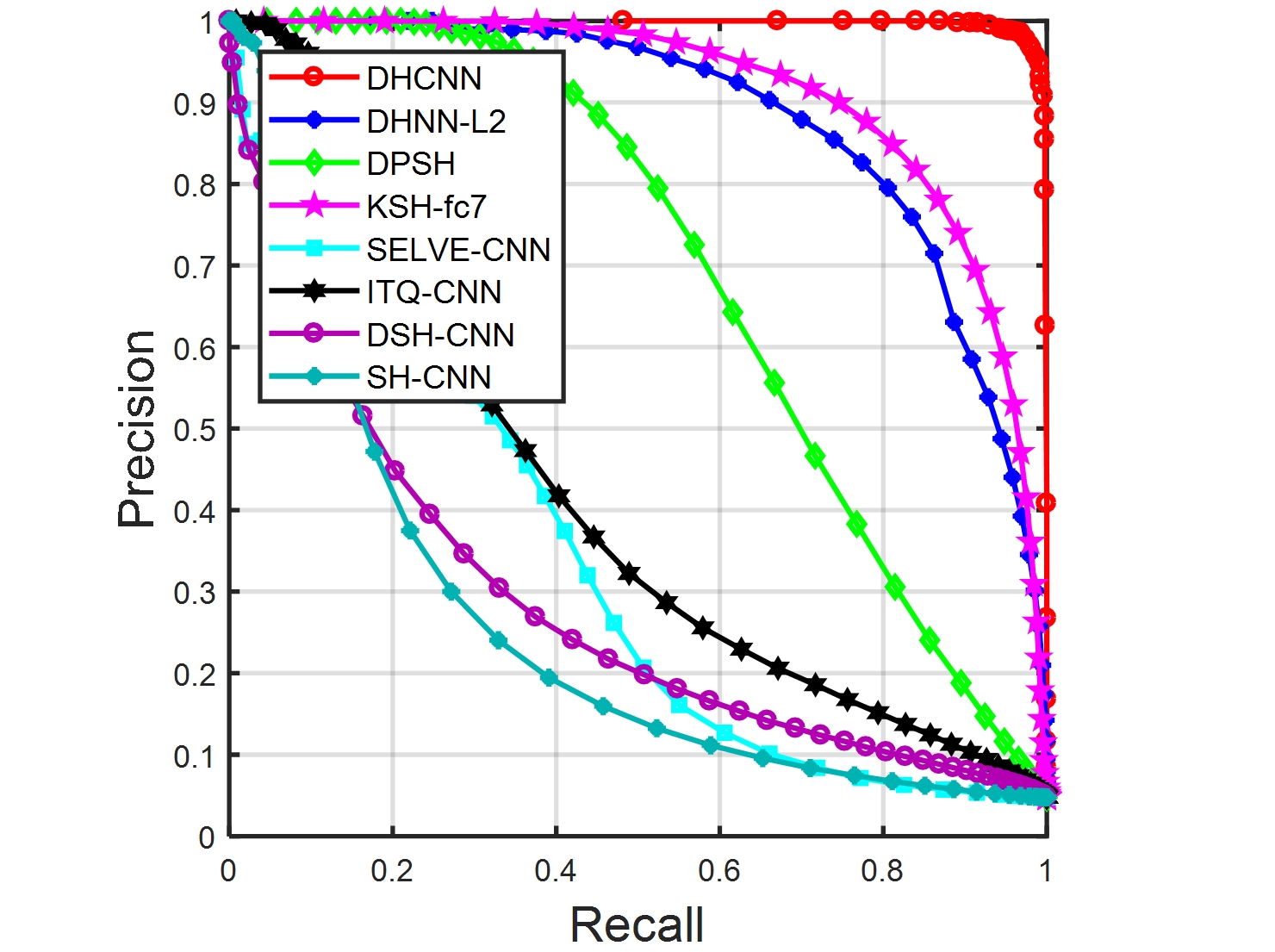}}
\caption{The retrieval results on UCMD with 64-bit hash code. (a) Recall@k; (b) Precision@k; (c) Precision-Recall}
\label{Curves-UCM21}
\end{center}
\end{figure*}

\begin{figure*}
\begin{center}
\subfigure[]{\includegraphics[width=57mm]{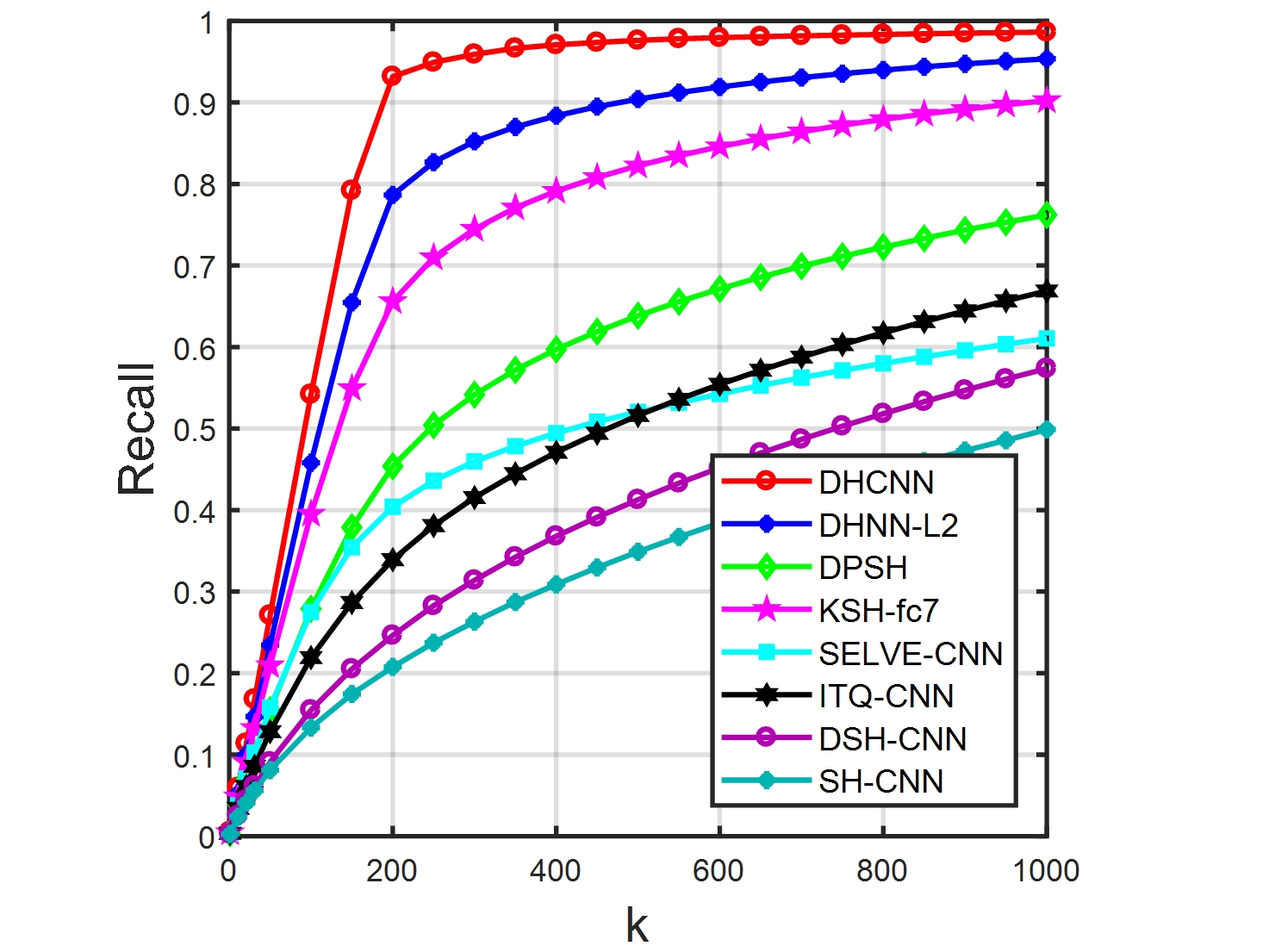}}
\subfigure[]{\includegraphics[width=57mm]{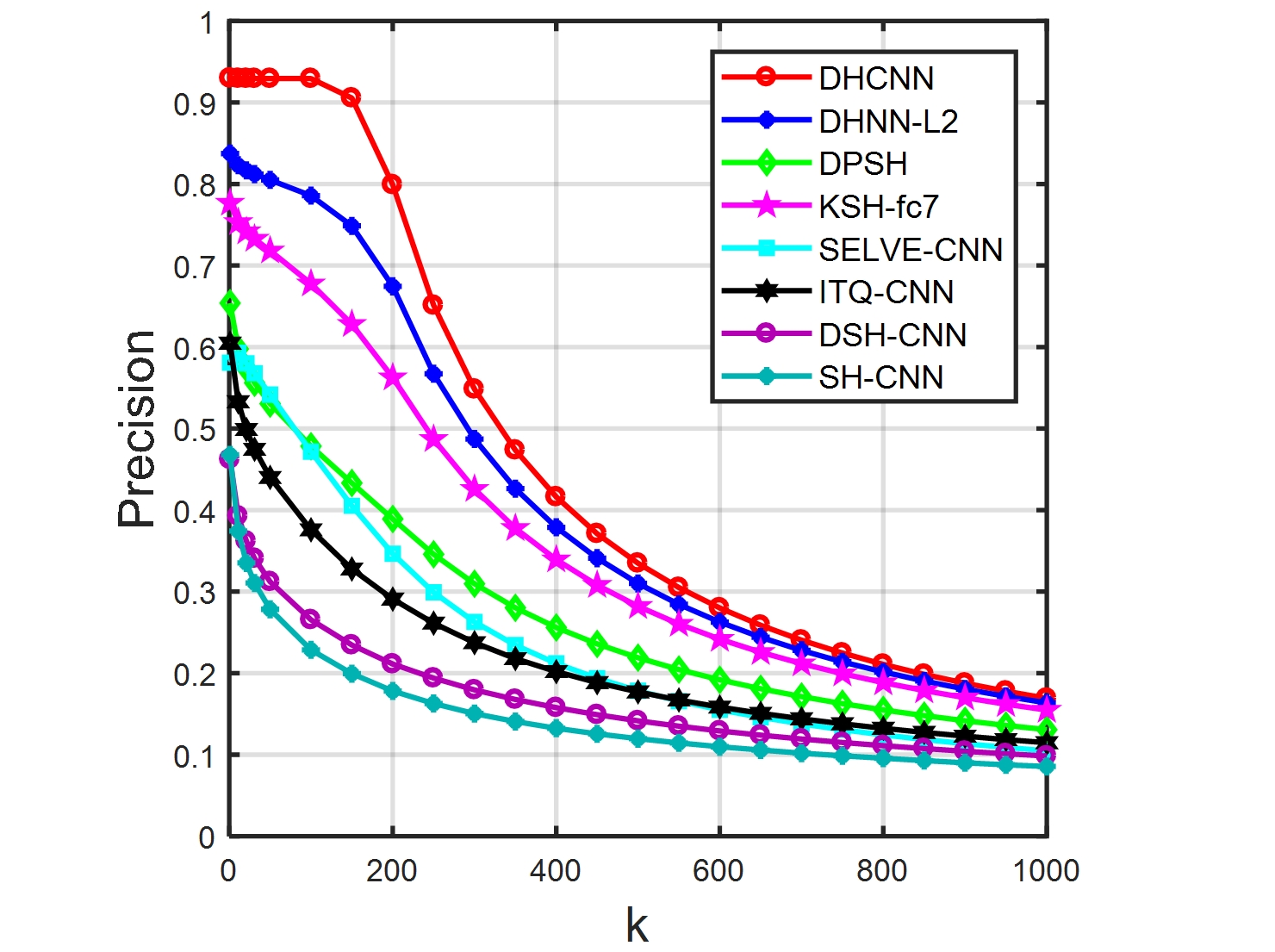}}
\subfigure[]{\includegraphics[width=55mm]{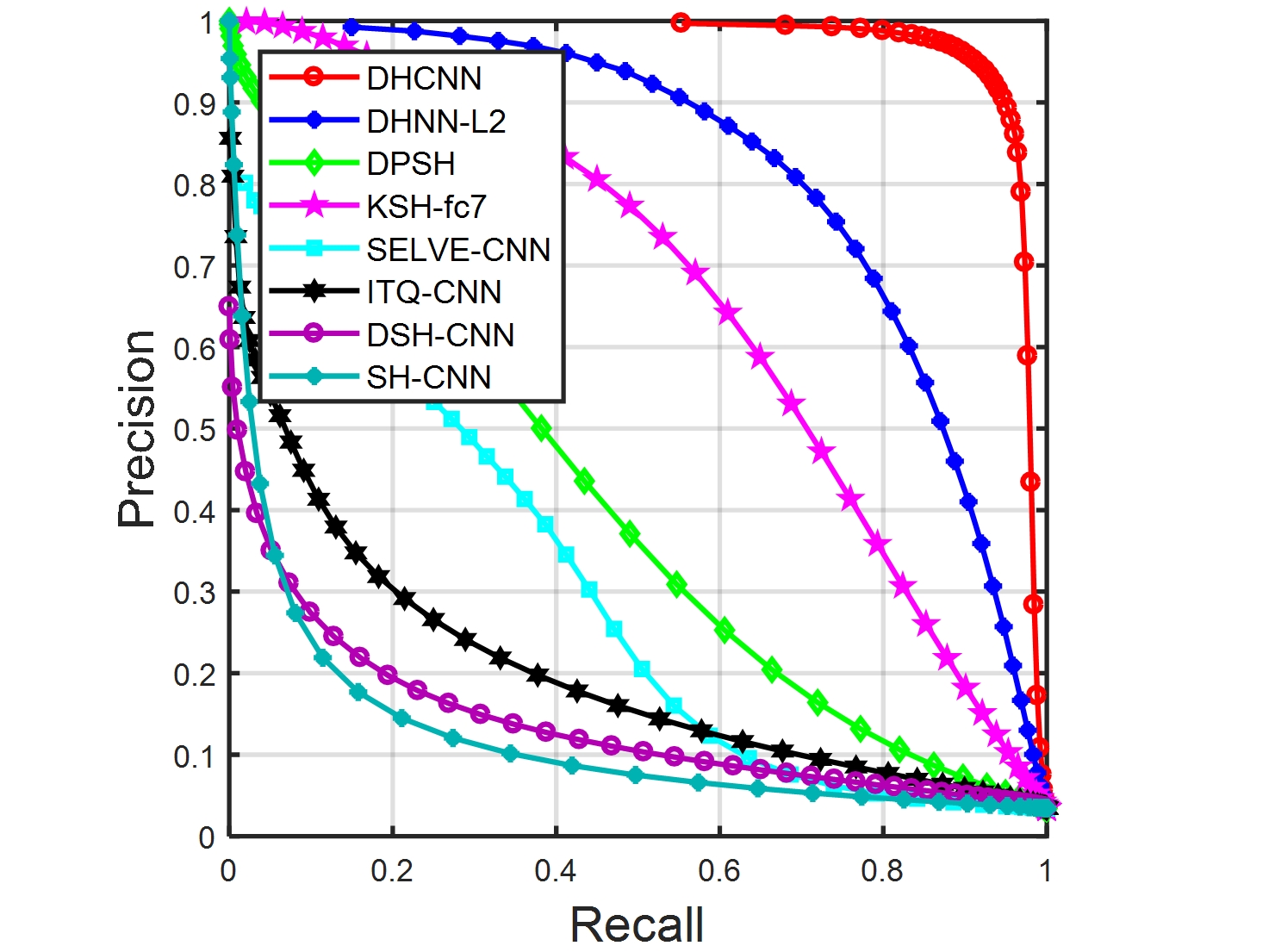}}
\caption{The retrieval results on AID with 64-bit hash code. (a) Recall@k. (b) Precision@k. (c) Precision-Recall}
\label{Curves-AID30}
\end{center}
\end{figure*}

In our experiments, we adopt four widely used metrics in the literature to evaluate the performance of retrieval methods.  \par
1) Mean Average Precision (MAP): In the query phase, we firstly rank all samples by computing the Hamming distance between the query sample and the database samples. Once obtaining the ranked list, we can get the average precision (AP) for each query image. Finally, the MAP can be computed via averaging the AP of all query images, which is defined as

\begin{equation}
\text{MAP}=\frac{1}{|Q|}\sum_{i=1}^{|Q|}\frac{1}{n_i}\sum_{j=1}^{n_i}P(i,j)
\end{equation}
where $|Q|$ is the volume of the query image set, $n_i$ is the number of images relevant to $i$th query image in the searching database, and $P(i,j)$ is the precision of the top $j$th retrieved
image of $i$th query image. \par

2) Precision@k: This metric measures precision value of the top $k$ retrieved images, which is defined as
\begin{equation}
\text{Precision@k} = \frac{n}{k}
\end{equation}
where $k$ and $n$ are the number of retrieved images and similar images to the query image in the top $k$ list, respectively.  \par

3) Recall@k: Recall@k computes the recall rate between the number of similar images to the query image in the top $k$ retrieved image and all similar images in database, which is defined as
\begin{equation}
\text{Recall@k} = \frac{n}{r}
\end{equation}
where $r$ and $n$ are the number of similar images in database and the top $k$ retrieved images, respectively.   \par

4) Precision-Recall: The Precision-Recall metric is another popular evaluation protocol in image retrieval, which plots the precision and recall rates at different searching Hamming radius. The first point of the Precision-Recall curve represents the precision and recall rate at Hamming radius equals to 0; the next point means the precision and recall rate at Hamming radius equals to 1, and so on. \par

The first evaluation metric, i.e., MAP, is used to evaluate the overall retrieval performance, the rest three metrics, including Precision@k, Recall@k, and Precision-Recall are used to compare the retrieval results of all methods in terms of Precision-k, Recall-k, and Precision-Recall curves, respectively.   \par

\subsection{Retrieval Results}

In our experiments, we run ten trials for each method and report averaged results. Considering that DPSH \cite{DPSH}, DHNN-L2 \cite{DHNN}, and our proposed DHCNN have utilized same framework to extract deep features, we firstly analyze the qualitative results for three methods on two datasets. Figs. \ref{UCM21-query} and \ref{AID30-query} show two query examples with top ten retrieved images on the UCMD and AID, respectively. For each query example, the top, middle, and bottom rows represent retrieval results obtained by DPSH \cite{DPSH}, DHNN-L2 \cite{DHNN}, and our proposed DHCNN, respectively. The green rectangle marks true positives, while the red rectangle marks false positives. The ``TL'' and ``PL'' represent the true label and predicted label of images, respectively. From Figs. \ref{UCM21-query} and \ref{AID30-query}, we can see that the three methods all achieve great retrieval performance for simple classes (e.g., airplane of UCMD and desert of AID) as shown in query 1. However, for some challenging classes which exhibit very high interclass similarity (e.g., medium residential and dense residential classes of UCMD, school and dense residential classes of AID), the DPSH \cite{DPSH} and DHNN-L2 \cite{DHNN} perform bad since several false positives are included in their retrieved images as shown in query 2. In contrast, our method still returns all true positives in top ten retrieved images, which exhibits the great advantage in coping with the problem of high interclass similarity in remote sensing images. More importantly, different from existing retrieval methods, our method can simultaneously retrieve the similar images and classify their semantic labels in a unified framework. As shown in Figs. \ref{UCM21-query} and \ref{AID30-query}, our method can precisely predict the true label of retrieved images and achieve satisfactory classification performance.    \par

We also report the quantitative comparison among all methods. Table \ref{Results-MAP} shows the image retrieval results in terms of MAP with different hash bits on the UCMD and AID. From this table, we can see that our proposed method significantly outperforms all the compared
state-of-the-art methods. In addition, we can also observe that the deep features-based methods (i.e., KSH-CNN \cite{KSH}, ITQ-CNN \cite{ITQ}, SELVE-CNN \cite{SELVE}, DSH-CNN \cite{DSH}, and SH-CNN \cite{SH}) largely outperform hand-crafted features-based methods (i.e., KSH-GIST \cite{KSH}, ITQ-GIST \cite{ITQ}, SELVE-GIST \cite{SELVE}, DSH-GIST \cite{DSH}, and SH-GIST \cite{SH}), which indicates deep features indeed exhibit great advantages over hand-crafted features for image feature representation. Although DPSH \cite{DPSH} and DHNN-L2 \cite{DHNN} have adopted same deep network as our method to learn hash codes for image retrieval, they only consider the similarity information between images. In contrast, our method utilizes similarity information between images and semantic information of each image to deliver the better performance. Specifically, the MAP of our method is about 33\% and 12\% higher than that of DPSH \cite{DPSH} and DHNN-L2 \cite{DHNN}, respectively, for 64-bit hash code on the UCMD. On the AID, the improved accuracies in terms of MAP achieve to be about 50\% and 17\% when compared with DPSH \cite{DPSH} and DHNN-L2 \cite{DHNN}, respectively, for 64-bit hash code.    \par

Finally, we further compare the retrieval results of different approaches in terms of Recall@k, Precision@k, and Precision-Recall metrics. In this experiment, the hand-crafted features-based methods (i.e., KSH-GIST \cite{KSH}, ITQ-GIST \cite{ITQ}, SELVE-GIST \cite{SELVE}, DSH-GIST \cite{DSH}, and SH-GIST \cite{SH}) are excluded mainly due to the very poor results obtained by these approaches. Figs. \ref{Curves-UCM21} and \ref{Curves-AID30} show the corresponding retrieval results on UCMD and AID, where the hash bit is set to 64 for all methods. From Figs. \ref{Curves-UCM21} and \ref{Curves-AID30}, we can see that our method outperforms all compared methods under different k for Recall@k and Precision@k metrics both on UCMD and AID. At the same time, the Precision-Recall curve also shows the great advantages of our method over other approaches on the two datasets.  \par

\subsection{Classification Results}

\begin{table}
\small
\centering
%\footnotesize
  \caption{ Classification Results in Terms of OA (Shown in Percentages) on UCMA and AID Datasets. The Ratio of Training Samples Is Set to 80\% and 50\% for UCMD and AID, Respectively. The Best Values Are Shown in Boldface.}
%\begin{tabular}{ccp{0.8cm}<{\centering}p{0.8cm}<{\centering}p{0.8cm}<{\centering}p{0.8cm}<{\centering}p{1cm}<{\centering}p{0.8cm}<{\centering}p{0.8cm}<{\centering}p{0.8cm}<{\centering}}
\begin{tabular}{ccc}
\hline
Method    & UCMD (80\%) & AID (50\%)  \\
\hline
\textbf{DHCNN} (Our method)  & \textbf{97.68}      & 93.48 \\
DCA \cite{DCA} & 96.90    & 89.72 \\
MSP-FV-AlexNet \cite{MSP} & 96.40  & 93.30 \\
MSP-FV-CaffeNet \cite{MSP} & 96.70  & 93.33 \\
MSP-FV-GoogLeNet \cite{MSP} & 92.60  & 85.90 \\
MSP-FV-VGG-F \cite{MSP} & 96.90  & 93.20 \\
MSP-FV-VGG-VD16 \cite{MSP}  & 97.60  & \textbf{94.10} \\
MSP-FV-VGG-VD19 \cite{MSP}  & 97.10  & 94.00 \\
CaffeNet \cite{AID} & 95.02  & 89.53 \\
GoogLeNet \cite{AID} & 94.32 & 86.39 \\
VGG-VD16 \cite{AID} & 95.21  & 89.64 \\

\hline
\end{tabular}
\label{Results-OA}
\end{table}

As mentioned above, the greatest advantage of our method over existing retrieval approaches is that our method can precisely classify the semantic label of the returned similar images. In other words, the retrieval and classification tasks can be simultaneously achieved in a unified framework. In this section, we conduct experiments on the UCMD and AID to validate the effectiveness of our method for remote sensing image classification. Specifically, the ratio of training samples is set to 80\% and 50\% per class for UCMD and AID, respectively. The hash bit is set to 64 for both datasets. We adopt overall accuracy (OA) as metric to evaluate the classification performance of all methods.   \par

In this experiment, we compare our method with some state-of-the-art methods for remote sensing image classification, including deep feature fusion based on discriminant correlation analysis (DCA) \cite{DCA}, multiscale pooling with Fisher vector method (MSP-FV) \cite{MSP}, and some deep-feature methods which extract the activations from the first fully connected layer of CaffeNet \cite{CaffeNet}, GoogLeNet \cite{GoogleNet}, and VGG-VD16 \cite{VGG}. Table \ref{Results-OA} shows the classification results obtained by all methods. From this table, we can see that our proposed method achieves great classification performance, i.e., 97.68\% and 93.48\% classification accuracies on UCMD and AID, respectively. It also can be seen that the OA of our method is slightly smaller than that of MSP-FV-VGG-VD16 \cite{MSP} on the AID. This experimental result is expectable since the adopted network model in MSP-FV-VGG-VD16 (i.e., VGG-VD16 \cite{VGG}) is much deeper than our model (i.e., VGG-F \cite{CNN-S}). In addition, we also observe that the GoogLeNet \cite{GoogleNet} performs worse than CaffeNet \cite{CaffeNet} and VGG-VD16 \cite{VGG}, which is inconsistent with image classification of natural images.   \par
%In one word, our method achieves satisfactory classification performance compared with recent state-of-the-art methods.  \par

\subsection{ Analysis of Parameters }

\begin{figure}[!tp]
\centering
\includegraphics[width=70mm]{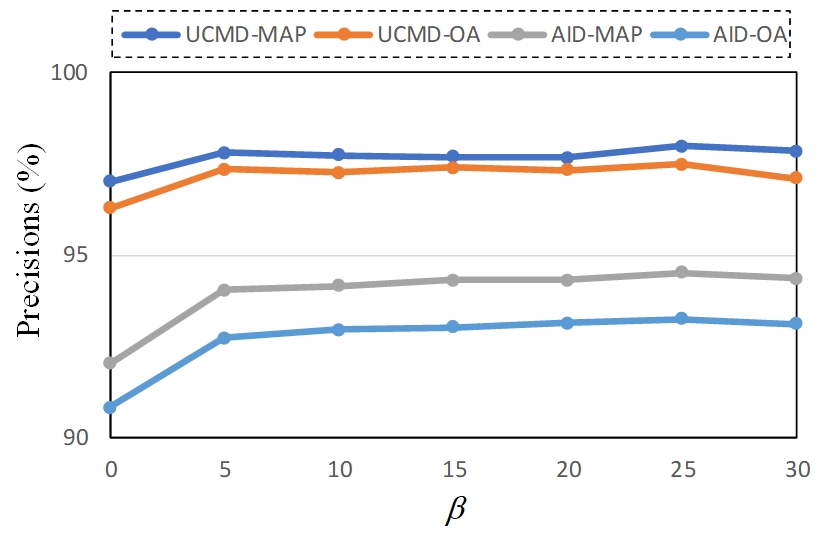}
\caption{Effect of $\beta$ on retrieval and classification precisions on the UCMD and AID.}
\label{beta}
\end{figure}

\begin{figure}[!tp]
\centering
\includegraphics[width=70mm]{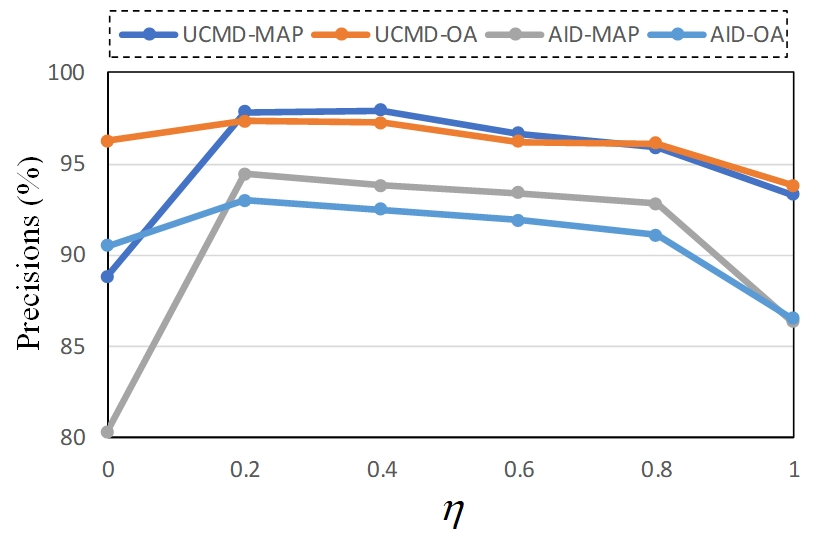}
\caption{Effect of $\eta$ on retrieval and classification precisions on the UCMD and AID.}
\label{eta}
\end{figure}

In our object function, parameter $\beta$ constrains hash-like feature $\mathbf{u}_i$ approach to hash code $\mathbf{b}_i$, parameter $\eta$ balances the semantic information and similarity information. In this section, the effects of $\beta$ and $\eta$ on retrieval and classification performance are analyzed.   \par
%In these experiments, the hash bit is set to 64 for two datasets.
Fig. \ref{beta} shows the effects of $\beta$ on MAP and OA on the UCMD and AID, where $\eta$ is set to 0.2 for two datasets. From Fig. \ref{beta}, we can see that the retrieval and classification accuracies are lowest when $\beta$ equals to 0. The main reason is that the learned hash codes lack of compactness since network cannot effectively constrain hash-like feature approach to hash code when $\beta$ equals to 0. The best retrieval and classification results are achieved when $\beta$ equals to 25 for both the UCMD and AID. In addition, Fig. \ref{eta} shows the effects of $\eta$ on MAP and OA on the UCMD and AID, where $\beta$ is set to 25 for two datasets. From this figure, we can see that when $\eta$ equals to 0 (i.e., only consider label loss of each image) and 1 (i.e., only consider similarity loss of between images), the MAP and OA significantly decrease for both UCMD and AID. The MAP and OA obtain the highest values when $\eta$ achieves to 0.2 for both the UCMD and AID. The above experimental results also validate the superiority of combining the semantic information of each image and similarity information between images.  \par

\section{Conclusion}

In this paper, we redefine the image retrieval problem as visual and semantic retrieval of images. Specifically, given a query image, a set of similar images and their semantic labels can be simultaneously obtained via a united framework. To this end, a novel DHCNN is proposed to learn compact hash codes for efficient RSIR and discriminative features for accurate semantic label
classification. In more detail, we first introduce a pre-trained CNN to extract high-dimensional deep features from raw remote sensing images. Then, a hash layer is perfectly inserted into the CNN to learn low-dimensional hash codes. In addition, a fully connected layer with a softmax function is performed on hash layer to generate class distribution. Finally, a loss function which simultaneously considers the semantic information and similarity information is elaborately designed to train DHCNN. The experimental results on UCMD and AID demonstrate that the proposed method gives excellent results as it achieves the state-of-art retrieval and classification performance.

\ifCLASSOPTIONcaptionsoff
  \newpage
\fi

\bibliographystyle{IEEEbib}
\bibliography{RSIR_RSIC_Refs}


\end{document}